\pdfoutput=1

\documentclass[11pt]{article}

\usepackage[preprint]{acl}

\usepackage{times}
\usepackage{latexsym}

\usepackage{algorithm}
\usepackage{algpseudocode}

\usepackage[most]{tcolorbox}
\definecolor{mygray}{rgb}{0.94,0.95,0.95}
\lstdefinelanguage{prompt}{
    basicstyle=\normalfont\fontfamily{pcr}\selectfont,
    showstringspaces=false,
    breaklines=True,
    backgroundcolor=\color{mygray},
}
\newtcolorbox[list inside=prompt,auto counter]{prompt}[1][]{
    colbacktitle=black!60,
    coltitle=white,
    fontupper=\footnotesize,
    boxsep=5pt,
    left=0pt,
    right=0pt,
    top=0pt,
    bottom=0pt,
    boxrule=1pt,
    #1,
}

\newcommand{\ours}[1]{\textbf{LeTS}}
\newcommand{\TODO}[1]{\textcolor{red}{\textbf{TODO}}}

\newcommand{\think}[1]{\textcolor{blue}{\texttt{<think>}} #1 \textcolor{blue}{\texttt{</think>}}}
\newcommand{\search}[1]{\textcolor{cyan}{\texttt{<search>}} #1 \textcolor{cyan}{\texttt{</search>}}}
\newcommand{\info}[1]{\textcolor{brown}{\texttt{<result>}} #1 \textcolor{brown}{\texttt{</result>}}}
\newcommand{\answer}[1]{\textcolor{purple}{\texttt{<answer>}} #1 \textcolor{purple}{\texttt{</answer>}}}

\usepackage{booktabs}
\usepackage{multirow}
\usepackage{graphicx}
\usepackage{amsmath}
\usepackage{amsfonts}
\usepackage{bbding}
\usepackage{makecell}
\usepackage{caption}
\usepackage{subcaption}
\usepackage{hyperref}
\usepackage{pifont}
\usepackage{CJKutf8}

\usepackage[T1]{fontenc}

\usepackage[utf8]{inputenc}

\usepackage{microtype}

\usepackage{inconsolata}

\usepackage{graphicx}

\title{\textit{LeTS}: Learning to Think-and-Search via Process-and-Outcome Reward Hybridization}

\author{
 \textbf{Qi Zhang\textsuperscript{1,2}$^*$},
 \textbf{Shouqing Yang\textsuperscript{1,2}$^*$},
 \textbf{Lirong Gao\textsuperscript{1}},
 \textbf{Hao chen\textsuperscript{1}},
\\
 \textbf{Xiaomeng Hu\textsuperscript{1}},
 \textbf{Jinglei Chen\textsuperscript{2}},
 \textbf{Jiexiang Wang\textsuperscript{2}},
 \textbf{Sheng Guo\textsuperscript{2}},
\\
 \textbf{Bo Zheng\textsuperscript{2}},
 \textbf{Haobo Wang\textsuperscript{1}$^\dagger$},
 \textbf{Junbo Zhao\textsuperscript{1}$^\dagger$}
\\
\\
 \textsuperscript{1}Zhejiang University,
 \textsuperscript{2}MYBank, Ant Group,
\\
 \texttt{\{cheung\_se,sqyang,gaolirong,h.c.chen,xm.hu,wanghaobo,j.zhao\}@zju.edu.cn},
\\
 \texttt{
  chenjinglei.cjl@mybank.cn,\{jxwang1994,guosheng1001\}@gmail.com,guangyuan@antgroup.com}
}

\begin{document}
\begin{CJK}{UTF8}{gbsn}
\maketitle

\footnotetext[1]{Work done during internship at MYBank, Ant Group}
\footnotetext[2]{Corresponding Author}

\begin{abstract}
    Large language models~(LLMs) have demonstrated impressive capabilities in reasoning with the emergence of reasoning models like OpenAI-o1 and DeepSeek-R1. Recent research focuses on integrating reasoning capabilities into the realm of retrieval-augmented generation~(RAG) via outcome-supervised reinforcement learning~(RL) approaches, while the correctness of intermediate think-and-search steps is usually neglected.
    To address this issue, we design a process-level reward module to mitigate the unawareness of intermediate reasoning steps in outcome-level supervision without additional annotation. Grounded on this, we propose \textbf{Le}arning to \textbf{T}hink-and-\textbf{S}earch~(\ours{}), a novel framework that hybridizes stepwise process reward and outcome-based reward to current RL methods for RAG.
    Extensive experiments demonstrate the generalization and inference efficiency of \ours{} across various RAG benchmarks. In addition, these results reveal the potential of process- and outcome-level reward hybridization in boosting LLMs' reasoning ability via RL under other scenarios. The code will be released soon.
\end{abstract}

\section{Introduction}
Large language models (LLMs)~\cite{qwen2.5, llama3, zhao2023survey} have exhibited remarkable performance across a wide range of downstream tasks, such as code generation~\cite{jiang2024survey}, mathematical reasoning~\cite{luo2023wizardmath}, and question answering~\cite{yue2025survey}.
However, relying solely on parametric knowledge stored within LLMs presents inherent limitations, such as generating hallucinated or outdated information~\cite{huang2025survey, haloscope}.
To mitigate these issues, integrating external knowledge sources through \emph{retrieval-augmented generation} (RAG) has become a widely adopted and effective paradigm~\cite{shuster2021retrieval, gao2023retrieval}. In particular, RAG empowers LLMs to dynamically access relevant information, thereby enhancing factual accuracy and adaptability in open-domain applications.


\begin{figure}[t]
  \centering
  \includegraphics[width=\columnwidth]{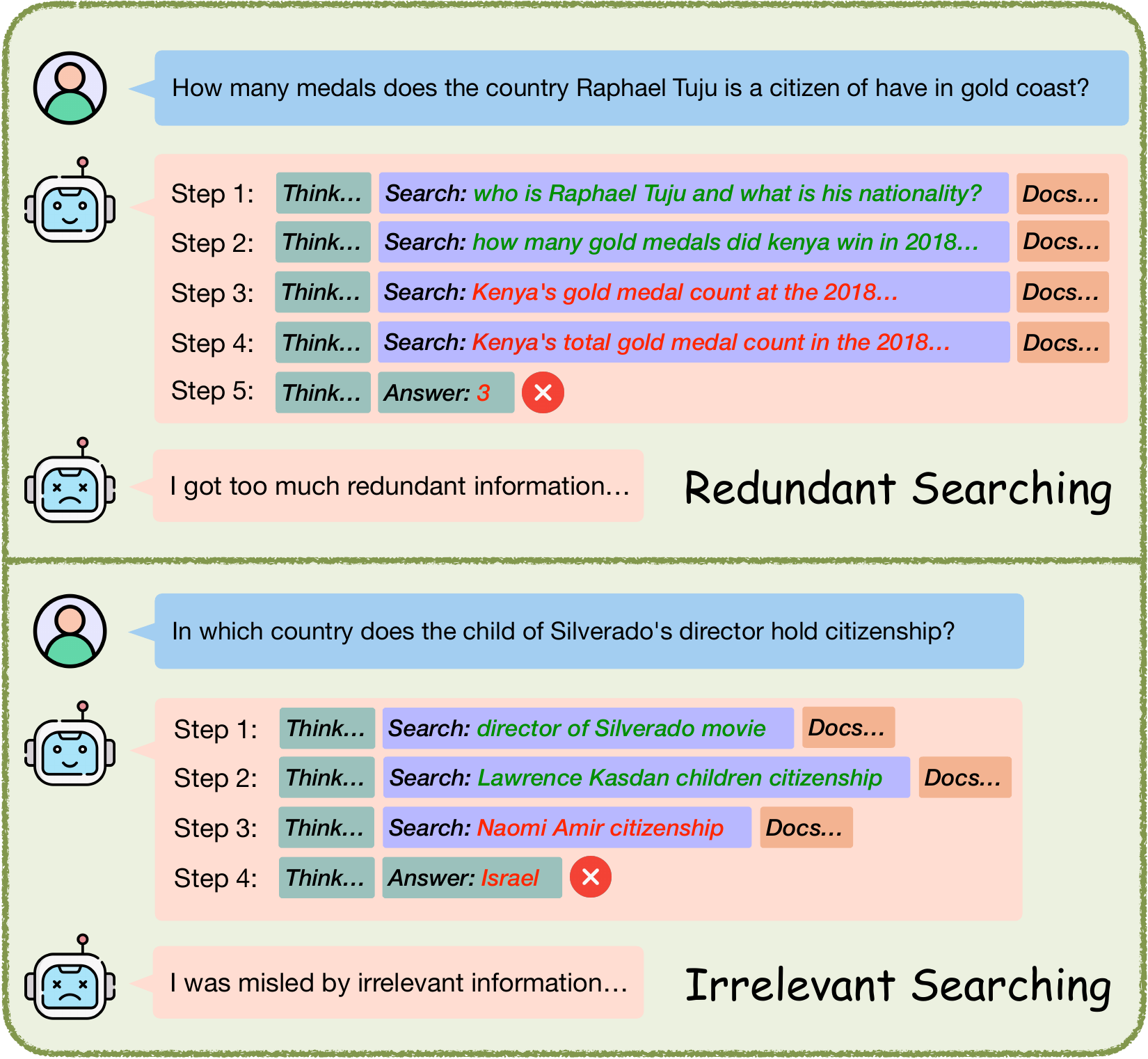}
  \caption{Illustrations of redundant searching issue and irrelevant searching issue in the current learning to think-and-search methods.}
  \label{fig:intro_fig}
\end{figure}

Despite its success, vanilla RAG methods struggle when faced with complicated or ambiguous queries~\cite{lewis2020retrieval, guu2020realm}, where a single-shot retrieval is often insufficient for multi-hop reasoning. To address this challenge, prompting-based methods have been proposed to equip LLMs with query decomposition capabilities by manually designing prompts~\cite{rqrag, ra-isf, ma2023query}. However, such approaches heavily depend on the inherent capacity of the language model and often fail to generalize across diverse scenarios. Alternatively, training-based approaches have attempted to distill this ability from larger models into smaller ones via supervised fine-tuning~\cite{selfrag, react}, yet they still suffer from limited generalization and lack robust reasoning supervision.


Building upon the success of large reasoning models such as OpenAI-o1~\cite{openai-o1} and DeepSeek-R1~\cite{deepseek-r1}, reinforcement learning (RL) techniques—particularly PPO and GRPO~\cite{deepseek-math}—have shown strong potential in enhancing the reasoning capabilities of LLMs. Motivated by this, recent studies have applied RL to develop the \emph{retrieving-when-needed} behavior in LLMs, often framed as a \emph{think-and-search} process~\cite{R1-searcher,Search-r1,ReSearch}. 
While these outcome-based methods have achieved promising results, they still struggle with two common issues that lead to suboptimal performance: \emph{i}) \textbf{Redundant Searching}, where the model repeatedly retrieves similar or unnecessary information across steps; and \emph{ii}) \textbf{Irrelevant Searching}, where irrelevant content is introduced during intermediate steps, ultimately misleading the model. These issues are empirically illustrated in Figure~\ref{fig:intro_fig}, and highlight the limitations of using coarse-grained, outcome-level rewards alone to supervise multi-step reasoning processes.

\begin{figure}[t]
  \centering
  \includegraphics[width=\columnwidth]{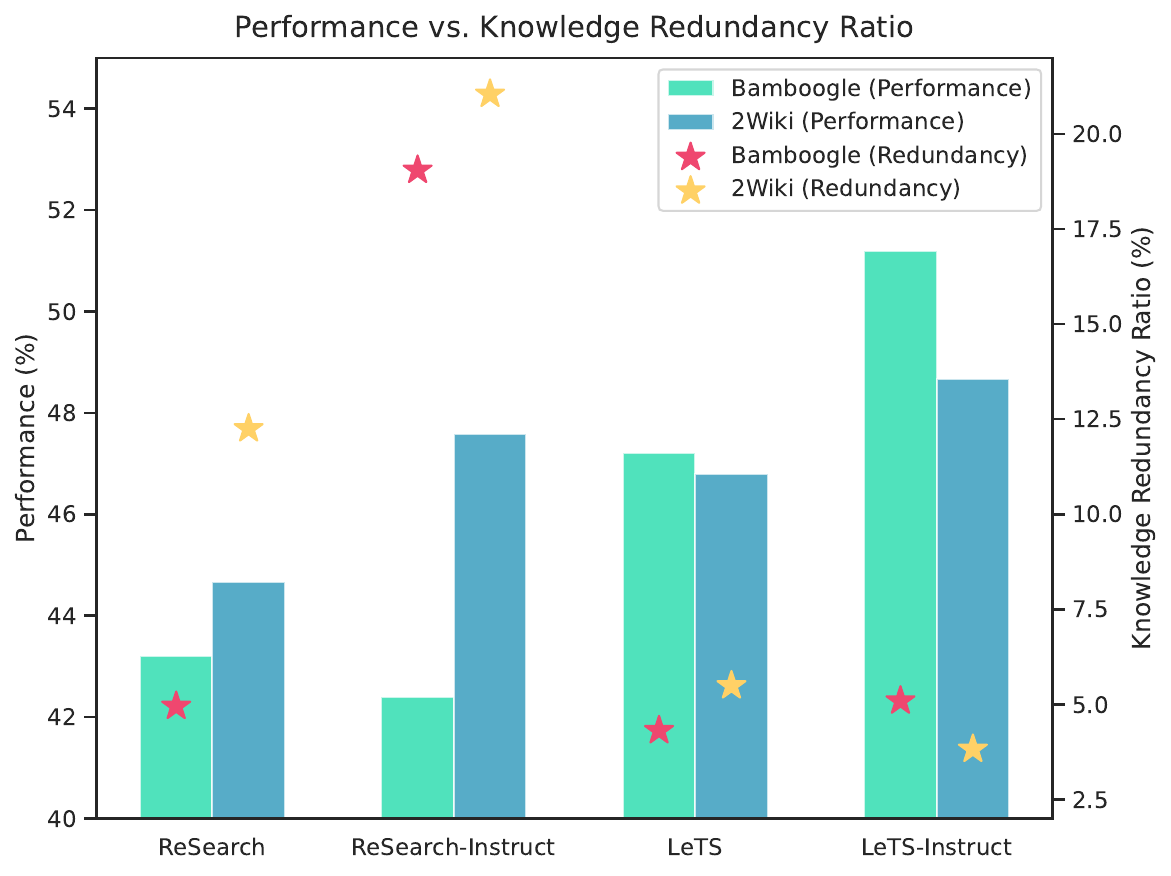}
  \caption{Collaborative comparison of performance and knowledge redundant frequency on two representative datasets with Qwen-7B-Base and Qwen-7B-Instruct as the backbone model.}
  \label{fig:performance_vs_redundancy}
\end{figure}

To better supervise the intermediate think-and-search steps, we introduce two complementary rule-based process-level reward modules to reform outcome-supervised GRPO approaches.
First, a knowledge redundancy reward penalizes steps that retrieve information already covered by superior rollouts.
Second, a knowledge match reward identifies and rewards correct actions in weaker rollouts by comparing them against high-performing rollouts.
Grounded on this, we further propose \ours{}, a novel RL framework that hybridizes process- and outcome-level reward via an advantage rescaling approach.  Notably, as illustrated in Figure~\ref{fig:performance_vs_redundancy}, \ours{} resolves the dilemma of knowledge redundancy and demonstrates strong performance on both base and instruct models. In general, through extensive experiments on numerous RAG benchmarks, \ours{} shows its effectiveness, strong generalization, as well as inference efficiency under various scenarios.

To sum up, our contributions can be concluded in threefold:
\begin{itemize}
    \item We propose a novel RL framework \ours{} for RAG, which first integrates rule-based process reward with outcome-level reward under the scenarios of RAG with reasoning.
    \item \ours{} effectively elicits the think-and-search ability of LLMs and achieves $\mathbf{2.61\%}$ performance gain across various RAG benchmarks. Moreover, the average number of generated tokens and search times are reduced by $\mathbf{11.15\%}$ and $\mathbf{30.85\%}$, respectively.
    \item The strong performance and inference efficiency of \ours{} highlight the potential of process- and outcome-level reward hybridization in boosting LLMs' reasoning ability via RL under other scenarios.
\end{itemize}

\begin{figure*}[h]
    \centering
    \includegraphics[width=\textwidth]{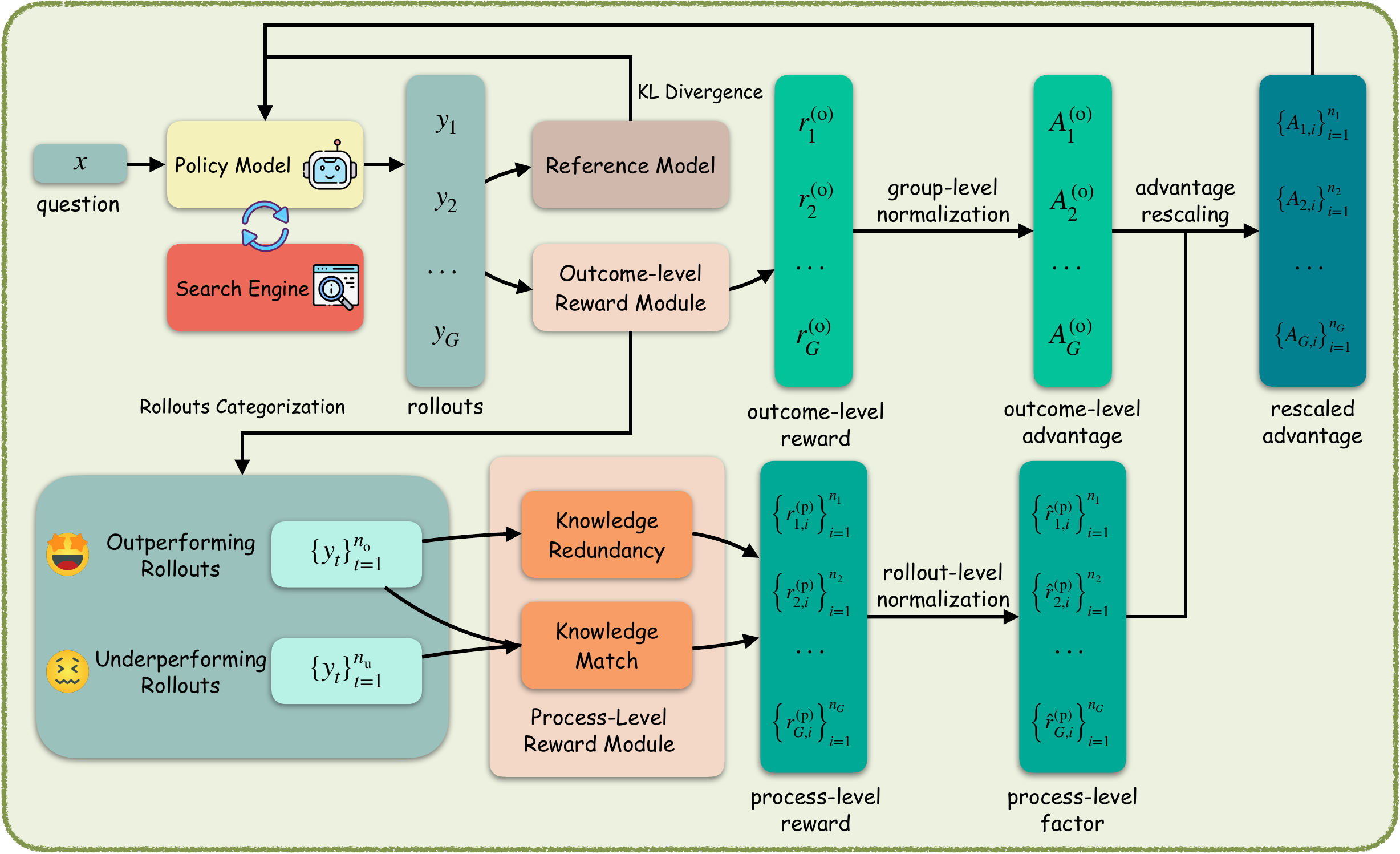}
    \caption{The core framework of \ours{} extends conventional outcome-supervised GRPO by introducing a fine-grained categorization of rollouts within each group. Specifically, generated rollouts are classified as either outperforming or underperforming, based on their relative returns. Separate reward functions are then applied to each category to compute process-level rewards. Finally, the rollout-level advantages are rescaled at the step level using an advantage normalization procedure, enabling more precise credit assignment during policy updates.}
    \label{fig:main}
\end{figure*}

\section{Related Work}
\subsection{Retrieval-Augmented Generation}
Retrieval-Augmented Generation (RAG) improves the factual accuracy of language models by incorporating external knowledge into the generation process. Early methods such as REALM~\cite{guu2020realm} and RAG~\cite{lewis2020retrieval} adopt a fixed retrieve-then-generate framework, which struggles with multi-hop reasoning due to static retrieval. To address this limitation, iterative approaches like IR-CoT~\cite{ircot} and Iter-RetGen~\cite{iter-retgen} retrieve evidence step-by-step using chain-of-thought prompting. While effective, these methods often depend on prompt engineering and large black-box LLMs. Recent work introduces learning-based retrieval policies to improve flexibility. ReSearch~\cite{ReSearch}, for instance, uses RL to determine when to retrieve. However, it relies solely on outcome-level rewards, ignoring the quality of intermediate reasoning steps.

\subsection{Large Language Models and Reinforcement Learning}
Reinforcement learning (RL) has been widely applied to teach LLMs retrieval-augmented reasoning. Prior methods such as ReSearch~\cite{ReSearch}, R1-Searcher~\cite{R1-searcher}, and Search-R1~\cite{Search-r1} rely on outcome-level rewards to guide retrieval behavior. However, they apply coarse-grained signals uniformly across reasoning steps, making it hard to distinguish informative from redundant actions. To address this, S$^2$R~\cite{S2R} introduces process-level supervision by training models to self-verify and self-correct using both process- and outcome-level rewards. While effective for structured reasoning tasks like math, its rewards are limited to verification quality and do not explicitly supervise retrieval behavior.

In contrast, our method designs rule-based, retrieval-specific process-level rewards, evaluating each think-and-search step for relevance, redundancy, and coherence. Combined with outcome-level feedback, this enables fine-grained control and more robust learning.

\section{Methodology}
In this section, we formalize the think-and-search task setting and introduce our proposed method, \ours{}, in detail. Specifically, \ours{} is primarily grounded in Group Relative Policy Optimization (GRPO), which estimates value baselines using a group of rollouts, in contrast to Proximal Policy Optimization (PPO) that relies on a separately trained critic model. The overall framework is illustrated in Figure~\ref{fig:main}, and a detailed pseudocode implementation is provided in Appendix~\ref{appendix:pseudo}.

\subsection{Think-and-Search Formulation}
\label{para:think-and-search}
Under our think-and-search scenarios, the language model $\mathcal{M}$ invokes the retriever $\mathcal{R}$ selectively, only when additional external information is deemed necessary. At each step, the retrieved documents are concatenated with the original context and used as input for the subsequent reasoning stage. For a given question $q$, the generation process with $n$ think-and-search steps is defined as:
\begin{equation}
    \begin{aligned}
        (t_i,s_i)&=\mathcal{M}\left(q,\{(t_j,s_j,d_j)\}_{j=1}^{i-1}\right) \\
        d_i&=\mathcal{R}\left(s_i\right)
\end{aligned}
\end{equation}, where $t_i$, $s_i$, and $d_i$ denote the thinking process, searching query, and retrieved documents of the $i$-th step, respectively. After completing $n$ such steps, the model produces the final reasoning trace $t_{n+1}$ and then generate the final answer $a$ as follow:
\begin{equation}
    (t_{n+1},a)=\mathcal{M}\left(q,\left\{(t_j,s_j,d_j)\right\}_{j=1}^{n}\right)
\end{equation}. A detailed instantiation of the think-and-search process is provided in Appendix~\ref{appendix:training_prompt}.

\subsection{Outcome-Level Reward Modeling}
Building on prior work in outcome-supervised RL~\cite{ReSearch,Search-r1,R1-searcher}, we define an outcome-level reward based on both the structural correctness of the reasoning format and the accuracy of the final answer. For format correctness, we verify whether the generated rollout adheres to the think-and-search reasoning chain described in Section~\ref{para:think-and-search}, as well as whether the final answer is properly structured. For answer correctness, we use the F1 score between the predicted answer and the gold reference, as exact match metrics are known to induce reward hacking behaviors~\citep{R1-searcher}. The outcome-level reward $r^\text{o}$ is then formulated as follow:
\begin{equation}
    r^\text{o}=
    \begin{cases}
    0.9\cdot\text{F1}(a_\text{pred},a_\text{gt})+0.1,&\mbox{correct format}\\
    0,&\mbox{wrong format}
    \end{cases}
\label{eq:outcome}
\end{equation}, where $\text{F1}(a_\text{pred},a_\text{gt})$ denotes the F1 score between the predicted answer $a_\text{pred}$ and the golden answer $a_\text{gt}$.

\subsection{Process-Level Reward Modeling}
Notably, relying solely on outcome-level rewards can lead to overlooking the correctness of intermediate reasoning steps (\emph{e.g.}, redundant operations in an otherwise successful rollout, or correct substeps within a failed one). To address this limitation, we introduce a process-level reward modeling framework tailored to our think-and-search RAG setting.

Based on the outcome-level reward defined in Equation~\ref{eq:outcome}, we categorize rollouts generated by the policy model during online RL into three types:
\emph{(i)} invalid rollouts that fail to conform to the required reasoning or answer format,
\emph{(ii)} outperforming rollouts that produce exact-match answers with correct format, and
\emph{(iii)} underperforming rollouts, which include all remaining cases that do not meet the above criteria.

Owing to the difficulty of reliably segmenting reasoning steps in malformed outputs, we omit the process-level reward for rollouts with the wrong format. For the latter two kinds of rollouts, we design two types of rule-based process-level reward: \textit{rollout-level knowledge redundancy reward} and \textit{group-level knowledge match reward} to address the two phenomena mentioned in Figure~\ref{fig:intro_fig}.

\paragraph{Rollout-Level Knowledge Redundancy Reward}
It's well-established that redundant knowledge in the RAG system can increase inference latency and even ruin the generation quality~\cite{ 10746232,chen2023understanding,tan-etal-2024-blinded}. To mitigate this issue, we introduce a rollout-level knowledge redundancy reward that penalizes reasoning steps that retrieve documents redundant with respect to earlier steps within the same rollout. To isolate the effect of the policy model from potential retriever bias, we compute pairwise redundancy only between the current and preceding steps. Specifically, we use the Jaccard score to quantify redundancy between two sets of retrieved documents. The process-level reward for the $j$-th step of the $i$-th rollout is thus defined as:
\begin{equation}
\label{eq:sim}
    r^\text{p}_{i,j}=
    1-\max\left\{\frac{|d_{i,j}\cap d_{i,t}|}{|d_{i,j}\cup d_{i,t}|}\right\}_{t=1}^{j-1}
\end{equation}, where $d_{i,j}$ represents the documents retrieved in the $j$-th step of the $i$-the rollouts. In this way, steps that introduce new external knowledge will be rewarded while the redundant ones will be punished.

\paragraph{Group-Level Knowledge Match Reward}
In contrast to outperforming rollouts, underperforming rollouts often suffer from irrelevant or inconsistent document retrieval~\citep{yoran2024making}. Inspired by self-consistency approaches used during inference~\citep{wang2023selfconsistency}, we propose a group-level knowledge match (KM) reward that leverages outperforming rollouts to supervise underperforming ones through a self-contrastive mechanism within RL. Given an outperforming rollout $y_o$ and an underperforming rollout $y_u$, we compute the document-level Jaccard similarity between each pair of steps $(\mathbf{s}_{u,j},\mathbf{s}_{o,k})$ from $y_u$ and $y_o$, respectively:
\begin{equation}
    \text{Match}\left(\mathbf{s}_{u,j},\mathbf{s}_{o,k}\right)=
    \frac{|d_{u,j}\cap d_{o,k}|}{|d_{u,j}\cup d_{o,k}|}
\label{eq:match}
\end{equation}
This yields a match matrix $\mathbf{m}^{u,o}\in\mathbb{R}^{n_u\times n_o}$ , where $n_u$ and $n_o$ denote the number of think-and-search steps in $y_u$ and $y_o$, respectively. To derive the optimal alignment between steps, we adopt the Kuhn-Munkres algorithm (also known as the Hungarian algorithm), which computes the maximum-weight bipartite matching between the two sets of steps. This naturally aligns with the framework of optimal transport~\cite{optimal_transport}, where steps in $y_u$ are optimally mapped to those in $y_o$ under a transport cost defined by their pairwise similarity. The resulting process-level reward vector for the underperforming rollout is given by:
\begin{equation}
    \left\{r^\text{p}_{u,j}\right\}_{j=1}^{n_u}=\text{Kuhn-Munkres}\left(\mathbf{m}^{u,o}\right)
\end{equation}, where each entry  ${m}^{u,o}_{j,k}$ is defined as $\text{Match}\left(\mathbf{s}_{u,j},\mathbf{s}_{o,k}\right)$. If multiple outperforming rollouts exist within the group, we select the one yielding the highest total process-level reward for alignment. If no valid outperforming rollout is available (\emph{i.e.}, no exact-match answer), the KM reward is omitted for that group.

\subsection{Process-and-Outcome Reward Hybridization}
While outcome-supervised RL offers strong performance with low annotation cost, it suffers from a coarse reward assignment: the outcome-level reward is uniformly applied across all tokens or steps within a rollout, thereby overlooking the relative importance of individual reasoning steps. To mitigate this limitation, we propose a fine-grained integration of process-level signals into the RL objective.

Specifically, we extend the GRPO framework by hybridizing outcome- and process-level rewards through a mechanism we call \textbf{stepwise outcome-level advantage rescaling}. This approach adjusts the outcome-level advantage at each reasoning step according to the corresponding process-level reward, enabling more precise advantage assignment and better alignment between learning signals and reasoning quality.

Following the conventional outcome-supervised GRPO framework, we first compute the outcome-level advantage for each rollout by normalizing its outcome reward within the group. Specifically, for the $i$-th rollout in a group of size $G$, the outcome-level advantage is given by: $A_i^\text{o}=\frac{r_i^\text{o}-\text{mean}\left(\left\{r_j^\text{o}\right\}_{j=1}^G\right)}{\text{std}\left(\left\{r_j^\text{o}\right\}_{j=1}^G\right)}$. The outcome-level advantage $A_i^\text{o}$ is then uniformly applied to all think-and-search steps within the rollout. In contrast, for the process-level reward, we adopt a local normalization strategy. Unlike process-supervised GRPO~\citep{deepseek-math}, which may rely on global reward signals, our process-level supervision is designed to capture the relative importance of each step within a single reasoning chain. Thus, for the $j$-th step in the $i$-th rollout, the normalized process-level reward is computed as: $\hat{r}_{i,j}^\text{p}=\frac{r_{i,j}^\text{p}-\text{mean}\left(\left\{r_{i,j}^\text{p}\right\}_{j=1}^{n_i}\right)}{\text{std}\left(\left\{r_{i,j}^\text{p}\right\}_{j=1}^{n_i}\right)}$, where $n_i$ denotes the number of think-and-search steps in rollout $i$. This formulation allows for fine-grained reward shaping within each rollout, facilitating more targeted advantage assignment across the reasoning trajectory.

Since the normalized process-level reward $\hat{r}^\text{p}$ captures only the relative importance of steps within a single rollout, we use it to modulate the global outcome-level advantage. Specifically, we apply a stepwise rescaling mechanism that adjusts the original advantage at each step based on its process-level signal. The rescaled advantage $A_{i,j}$ for the $j$-th step in the $i$-th rollout is defined as:
\begin{equation}
    A_{i,j}=\left(1+\operatorname{sgn}\left(A_{i}^\text{o}\right)\cdot\lambda\hat{r}_{i,j}^\text{p}\right)A_{i}^\text{o}
\label{eq:rescale}
\end{equation}, where function $\operatorname{sgn}(A_{i}^\text{o})$ denotes the sign of the outcome-level advantage and $\lambda$ is a scaling factor controlling the magnitude of rescaling. This formulation amplifies or attenuates the original advantage based on stepwise quality, while preserving the sign of the overall trajectory’s reward signal.

For robustness, we do not apply rescaling to the final reasoning step and to rollouts excluded from process-level reward computation (\emph{e.g.}, those with invalid formats), ensuring consistent gradient propagation in these cases.

\begin{table*}[h]
\centering
\resizebox{\textwidth}{!}{%
\begin{tabular}{@{}lcccccccccc@{}}
\toprule[1.5pt]
\multicolumn{1}{c}{\multirow{2}{*}{Model}} & \multicolumn{2}{c}{HotpotQA}       & \multicolumn{2}{c}{2Wiki}          & \multicolumn{2}{c}{MusiQue}        & \multicolumn{2}{c}{Bamboogle}      & \multicolumn{2}{c}{Average} \\ \cmidrule(l){2-11} 
\multicolumn{1}{c}{}                       & \emph{EM}    & \emph{LJ}                         & \emph{EM}    & \emph{LJ}                         & \emph{EM}    & \emph{LJ}                         & \emph{EM}    & \emph{LJ}                         & \emph{EM}           & \emph{LJ}           \\ \midrule
\multicolumn{11}{c}{\textit{Qwen2.5-7B-Base/Instruct}}                                                                                                                                                                       \\ \midrule
Direct                           & 19.18 & \multicolumn{1}{c|}{30.64} & 25.76 & \multicolumn{1}{c|}{27.87} & 3.76  & \multicolumn{1}{c|}{10.38} & 10.40 & \multicolumn{1}{c|}{22.40} & 14.78        & 22.82        \\
Naïve RAG                                  & 31.90 & \multicolumn{1}{c|}{49.59} & 25.78 & \multicolumn{1}{c|}{29.52} & 6.21  & \multicolumn{1}{c|}{12.78} & 20.80 & \multicolumn{1}{c|}{32.00} & 21.17        & 30.97        \\
Iter-RetGen                                & 34.36 & \multicolumn{1}{c|}{52.22} & 27.92 & \multicolumn{1}{c|}{31.86} & 8.69  & \multicolumn{1}{c|}{16.14} & 21.60 & \multicolumn{1}{c|}{35.20} & 23.14        & 33.86        \\
IR-CoT                                     & 30.33 & \multicolumn{1}{c|}{52.06} & 21.57 & \multicolumn{1}{c|}{30.65} & 6.99  & \multicolumn{1}{c|}{14.19} & 24.80 & \multicolumn{1}{c|}{36.80} & 20.92        & 33.43        \\
ReSearch-Qwen-7B                           & 40.57 & \multicolumn{1}{c|}{60.26} & 44.67 & \multicolumn{1}{c|}{50.06} & 21.68 & \multicolumn{1}{c|}{32.19} & 43.20 & \multicolumn{1}{c|}{54.40} & 37.53        & 49.23        \\
ReSearch-Qwen-7B-Instruct                  & \textbf{43.52} & \multicolumn{1}{c|}{\underline{63.62}} & \underline{47.59} & \multicolumn{1}{c|}{\underline{54.22}} & \underline{22.30} & \multicolumn{1}{c|}{33.43} & 42.40 & \multicolumn{1}{c|}{54.40} & 38.95        & \underline{51.42}        \\ \midrule
\textbf{LeTS-Qwen-7B} & 40.62 & \multicolumn{1}{c|}{61.00} & 46.79 & \multicolumn{1}{c|}{52.21} & 22.18 & \multicolumn{1}{c|}{\textbf{34.09}} & \underline{47.20} & \multicolumn{1}{c|}{\underline{57.60}} & \underline{39.20} & 51.22 \\
\textbf{LeTS-Qwen-7B-Instruct} & \underline{43.16} & \multicolumn{1}{c|}{\textbf{63.69}} & \textbf{48.66} & \multicolumn{1}{c|}{\textbf{55.26}} & \textbf{23.21} & \multicolumn{1}{c|}{\underline{33.64}} & \textbf{51.20} & \multicolumn{1}{c|}{\textbf{62.40}} & \textbf{41.56} & \textbf{53.75} \\ \midrule
\multicolumn{11}{c}{\textit{Qwen2.5-3B-Instruct}}                                                                                                                                                                       \\ \midrule
Direct                           & 16.00 & \multicolumn{1}{c|}{24.56}      & 24.81 & \multicolumn{1}{c|}{26.47}      & 2.28  & \multicolumn{1}{c|}{6.04}      & 2.40  & \multicolumn{1}{c|}{7.20}      & 11.37        & 16.07              \\
Naïve RAG                                  & 27.97 & \multicolumn{1}{c|}{42.61}      & 24.38 & \multicolumn{1}{c|}{27.46}      & 5.75  & \multicolumn{1}{c|}{10.05}      & 9.60  & \multicolumn{1}{c|}{17.60}      & 16.93        & 24.43              \\
Iter-RetGen                                & 30.05 & \multicolumn{1}{c|}{45.17}      & 26.10 & \multicolumn{1}{c|}{29.35}      & 7.12  & \multicolumn{1}{c|}{12.04}      & 12.00 & \multicolumn{1}{c|}{17.60}      & 18.82        & 26.04              \\
IR-CoT                                     & 27.48 & \multicolumn{1}{c|}{50.67}      & 21.66 & \multicolumn{1}{c|}{34.14}      & 6.74  & \multicolumn{1}{c|}{14.78}      & 22.40 & \multicolumn{1}{c|}{33.60}      & 19.57        & 33.30             \\
ReSearch-Qwen-3B-Instruct                  & \underline{35.62} & \multicolumn{1}{c|}{\underline{54.40}}      & \underline{39.26} & \multicolumn{1}{c|}{\underline{46.10}}      & \underline{17.34} & \multicolumn{1}{c|}{\textbf{27.10}}      & \underline{37.60} & \multicolumn{1}{c|}{48.80}      & \underline{32.45} & \underline{44.10}             \\ \midrule
\textbf{LeTS-Qwen-3B-Instruct} & \textbf{37.06} & \multicolumn{1}{c|}{\textbf{55.21}}      & \textbf{41.04} & \multicolumn{1}{c|}{\textbf{47.48}} & \textbf{17.50} & \multicolumn{1}{c|}{\underline{26.93}}      & \textbf{38.40} & \multicolumn{1}{c|}{\textbf{51.20}} & \textbf{33.50} & \textbf{45.21}             \\ \bottomrule[1.5pt]
\end{tabular}%
}
\caption{Performance of our method compared to other methods with Qwen2.5-7B-Base/Instruct and Qwen2.5-3B-Instruct as the backbone model on four multi-hop benchmarks. The best results are highlighted in \textbf{bold}. Specifically, \emph{EM} and \emph{LJ} refer to Exact Match and LLM-as-a-Judge, respectively.}
\label{tab:main_results}
\end{table*}

\subsection{Reinforcement Learning Formulation}
We now formalize the overall training objective of \ours{}. Given a policy model $\pi_{\theta_\text{old}}$ and a reference model $\pi_{\theta_\text{ref}}$, based on $G$ rollouts $\{y_i\}_{i=1}^G\sim\pi_{\theta_\text{old}}(\cdot|x)$ for input $x\sim \mathcal{D}$, the objective of \ours{} is to optimize the policy $\pi_{\theta}$ by maximizing the following objective:
\begin{equation}
    \begin{split}
        &\mathcal{J}(\theta)=\mathbb{E}_{x\sim D,\{y_i\}_{i=1}^G\sim\pi_{\theta_\text{old}}(\cdot|x)}\frac{1}{G}\sum_{i=1}^{G}\frac{1}{|y_i|}\sum_{t=1}^{|y_i|}\\
        &[\min(\frac{\pi_{\theta}(y_{i,t}|x,y_{i,<t})}{\pi_\text{old}(y_{i,t}|x,y_{i,<t})}\hat{A}_{i,t},\text{clip}(\frac{\pi_{\theta}(y_{i,t}|x,y_{i,<t})}{\pi_\text{old}(y_{i,t}|x,y_{i,<t})},\\
        &1-\epsilon,1+\epsilon)\hat{A}_{i,t})-\beta\mathbb{D}_\text{KL}(\pi_{\theta}|\pi_\text{ref})]
    \end{split}
\label{eq:grpo}
\end{equation}, where $\hat{A}_{i,t}=A_{i,j(t)}$ is the rescaled advantage for the step that token $y_{i,t}$ belongs to, and $\epsilon$ is the clipping threshold. The mapping $j(t)$ associates each token position $t$ with its corresponding think-and-search step index within the rollout. The KL divergence term regularizes the policy to remain close to the reference model, weighted by the coefficient $\beta$. Following common practice~\citep{Search-r1,ReSearch,R1-searcher}, we mask out retrieved tokens during training to prevent noise in the retrieval content from skewing the learning signal.

\section{Experiments}
\subsection{Experimental Settings}
\paragraph{Datasets}

For training \ours{}, we largely adopt the experimental setup of ReSearch~\cite{ReSearch} for consistency and reproducibility. Specifically, we use the training split of MusiQue~\cite{trivedi-etal-2022-musique}, as it contains diverse types of multi-hop questions and has been constructed through rigorous quality control procedures. As for the evaluation benchmarks and metrics, please refer to~\ref{appendix:exp}

\paragraph{Baselines}
We compare \ours{} against several representative baselines:  
(1) \textbf{Direct} uses the instruction-tuned model to generate answers without any retrieval.  
(2) \textbf{Naïve RAG} appends retrieved documents to the input query and generates answers in a single forward pass.  
(3) \textbf{Iter-RetGen}~\cite{iter-retgen} alternates between retrieval and generation in a step-by-step manner.  
(4) \textbf{IR-CoT}~\cite{ircot} interleaves retrieval with chain-of-thought prompting to support multi-hop reasoning.  
(5) \textbf{ReSearch}~\cite{ReSearch} applies outcome-level RL to optimize retrieval strategies.  
We adopt ReSearch as our primary strong baseline due to its strong empirical performance and publicly available implementation. Other comparable methods, such as \textbf{Search-R1}~\cite{Search-r1} and \textbf{R1-Searcher}~\cite{R1-searcher} follow similar paradigms and are therefore subsumed under ReSearch in our evaluation.


\paragraph{Implementation Details}
We conduct our experiments and evaluations on Qwen2.5-7B, Qwen2.5-7B-Instruct, and Qwen2.5-3B-Instruct. For the codebase, we utilize two widely adopted frameworks: \texttt{VeRL}\footnote{\url{https://github.com/volcengine/verl}} for RL training, and \texttt{FlashRAG}\footnote{\url{https://github.com/RUC-NLPIR/FlashRAG}} for evaluation. All models are trained on the MusiQue training set for 2 epochs.  

For the rescaling factor $\lambda$, we perform hyperparameter tuning over \{0.05, 0.1, 0.2\} using Qwen2.5-7B-Instruct, and set $\lambda = 0.1$ for all experiments in this paper. Additional implementation details of \ours{} and the hyperparameter configurations for GRPO can be found in Appendix~\ref{appendix:implementation_details}.

\subsection{Main Results}
The main results of our method, compared against other baselines on Qwen2.5-7B-Base/Instruct and Qwen2.5-3B-Instruct, are presented in Table~\ref{tab:main_results}. Overall, our proposed \ours{} consistently outperforms all baselines, demonstrating its effectiveness in equipping LLMs with think-and-search capabilities. Specifically, LeTS-Qwen-7B achieves average improvements of $1.67\%$ and $1.99\%$, while LeTS-Qwen-7B-Instruct achieves $2.61\%$ and $2.33\%$ over previous RL-based methods using the same backbone, measured by \emph{EM} and \emph{LJ}, respectively.

Notably, \ours{} trained on Qwen2.5-7B-Base achieves performance comparable to methods trained on Qwen2.5-7B-Instruct. In addition, our method performs well on smaller models (\emph{i.e.}, Qwen2.5-3B-Instruct), indicating that instruct-tuned models of smaller scale can also benefit from our proposed hybrid reward framework. We exclude Qwen2.5-3B-Base from our analysis, as reinforcement learning on this model quickly degenerates due to reward hacking, resulting in behavior similar to vanilla R1-style methods without retrieval. Further details are provided in Appendix~\ref{appendix:slm}.

\subsection{Generalization on Single-Hop Benchmarks}
To further evaluate the generalization ability of \ours{} in single-hop scenarios, we conduct experiments on three widely used single-hop benchmarks: NQ, PopQA, and TriviaQA. As shown in Table~\ref{tab:single-hop}, \ours{} continues to outperform other baselines, demonstrating its effectiveness beyond multi-hop settings.

\begin{table}[h]
\centering
\resizebox{\columnwidth}{!}{%
\begin{tabular}{@{}l|ccc|c@{}}
\toprule[1.5pt]
\multicolumn{1}{c|}{Method} & \multicolumn{1}{c}{NQ} & \multicolumn{1}{c}{PopQA} & \multicolumn{1}{c|}{TriviaQA} & \multicolumn{1}{c}{Average} \\ \midrule
Direct & 14.24 & 15.11 & 43.60 & 24.32 \\
Naïve RAG        & 36.26 & 39.93 & 61.24 & 45.81 \\
Iter-RetGen      & 38.59 & 42.78 & 62.90 & 48.09 \\
IR-CoT           & 23.24 & 34.44 & 53.97 & 37.22 \\
ReSearch         & 40.86 & \textbf{44.58} & 63.71 & 49.72 \\ \midrule
\ours{}          & \textbf{42.58} & 43.91 & \textbf{64.63} & \textbf{50.37}\\ \bottomrule[1.5pt]
\end{tabular}%
}
\caption{Performance comparison between \ours{} with other baselines on single-hop benchmarks with Qwen-7B-Instruct as the backbone model.}
\label{tab:single-hop}
\end{table}

\section{Ablation Study and Analysis}
In this section, we present ablation studies and detailed analytical experiments to further investigate the performance of \ours{}. Unless otherwise stated, all experiments are conducted using Qwen2.5-7B-Instruct.

\subsection{Impact of Process-Level Reward}
We conduct ablation experiments to examine the effectiveness of the two types of process-level rewards. As shown in Table~\ref{tab:ablation}, both the rollout-level knowledge redundancy reward and the group-level knowledge match reward significantly reduce the model’s average search time while maintaining strong overall performance.

\begin{table}[h]
\centering
\resizebox{0.9\columnwidth}{!}{%
\begin{tabular}{@{}l|ccccc@{}}
\toprule[1.5pt]
\multicolumn{1}{c|}{\multirow{2}{*}{Method}} &
  \multicolumn{2}{c}{Settings} &
  \multicolumn{1}{c}{\multirow{2}{*}{\emph{AST}$\downarrow$}} &
  \multirow{2}{*}{\emph{EM}$\uparrow$} \\ \cmidrule(lr){2-3}
\multicolumn{1}{c|}{} &
  \multicolumn{1}{c}{w/. CR} &
  \multicolumn{1}{c}{w/. CM} &
  \multicolumn{1}{c}{} &
   &
  \multicolumn{1}{c}{} \\ \midrule
vanilla GRPO & \XSolidBrush & \XSolidBrush & 3.21 & 22.18 \\
w/. KR       & \Checkmark & \XSolidBrush & 2.71 & 22.22 \\
w/. KM       & \XSolidBrush & \Checkmark & 2.59 & 22.67 \\
\ours{}         & \Checkmark & \Checkmark & \textbf{2.21} & \textbf{23.21} \\ \bottomrule[1.5pt]
\end{tabular}%
}
\caption{The ablation on two types of process-level reward on the MusiQue dataset. \emph{AST} stands for the average search time for each method across all data points. To be specific, vanilla GRPO here refers to the GRPO algorithm that solely relies on outcome-level supervision.}
\label{tab:ablation}
\end{table}

\begin{table*}[]
\centering
\resizebox{0.9\textwidth}{!}{%
\begin{tabular}{@{}cc|cccc|c@{}}
\toprule
\multicolumn{2}{c|}{Datasets}             & HotpotQA & 2WikiMultiHopQA  & MusiQue & Bamboogle & Average \\ \midrule
\multirow{3}{*}{Num. of Tokens}       & ReSearch & 278.25   & 328.53 & 328.14  & 232.87    & 291.95  \\
                               & LeTS     & 244.29   & 275.82 & 335.87  & 189.30    & 261.32  \\
                               & $\Delta(\%)$ & $\downarrow$12.20    & $\downarrow$16.04  & $\uparrow$2.36   & $\downarrow$18.71     & $\downarrow$11.15   \\ \midrule
\multirow{3}{*}{Num. of Search Times} & ReSearch & 2.78     & 3.48   & 3.21    & 2.58      & 3.01    \\
                               & LeTS     & 1.84     & 2.31   & 2.21    & 1.94      & 2.08    \\
                               & $\Delta(\%)$ & $\downarrow$33.81    & $\downarrow$33.62  & $\downarrow$31.15   & $\downarrow$24.81     & $\downarrow$30.85   \\ \bottomrule
\end{tabular}%
}
\caption{Average number of tokens and search times of \ours{} comparing to ReSearch. Notably, tokens from the retrieved documents are not counted here, since they are not generated by the language models.}
\label{tab:efficiency}
\end{table*}

\begin{figure}[h]
  \centering
  \includegraphics[width=\columnwidth]{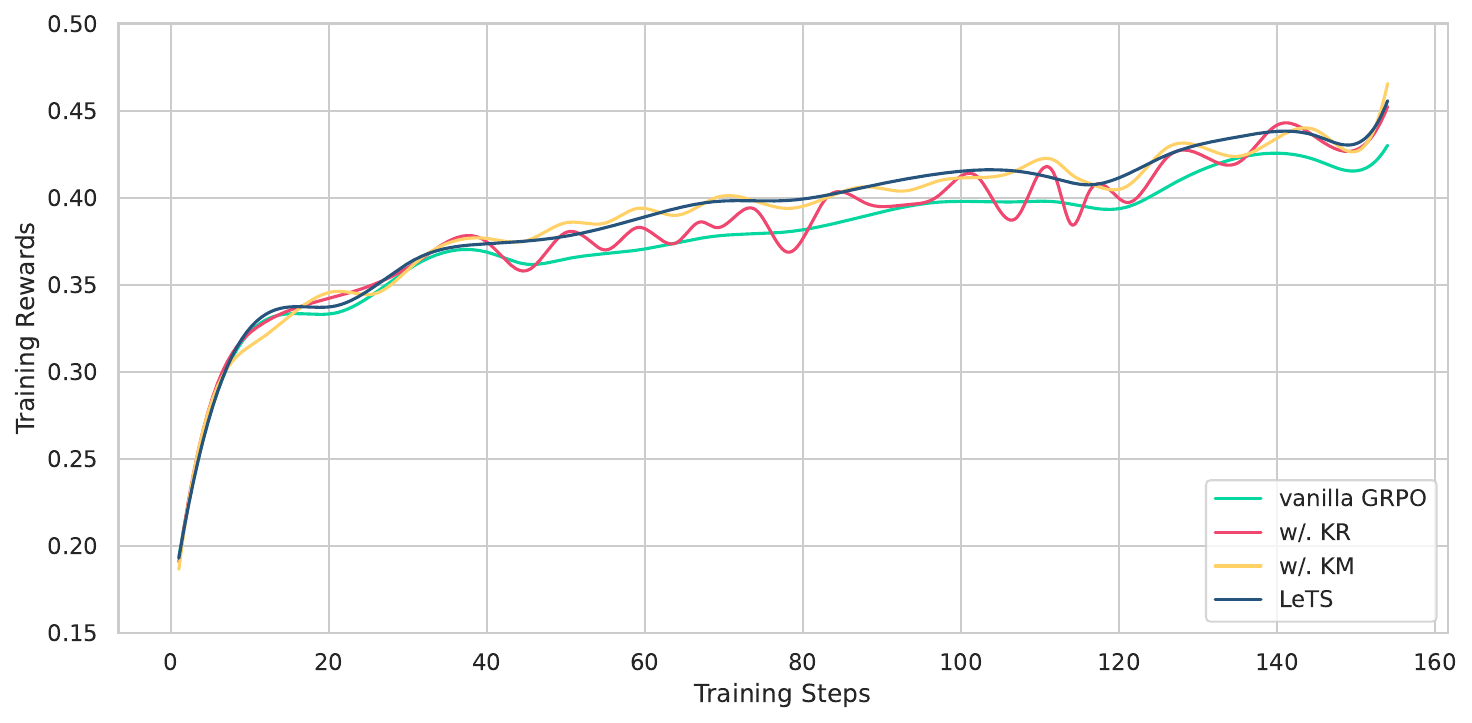}
  \caption{Training reward during training for the four methods.}
  \label{fig:training_rewards}
\end{figure}

Moreover, Figure~\ref{fig:training_rewards} illustrates the training dynamics of both the reward signals and knowledge redundancy. The training reward improves with the inclusion of either process-level component, and their combined usage leads to a more stable and consistent reward trajectory during RL training.

\begin{figure}[h]
  \centering
  \includegraphics[width=\columnwidth]{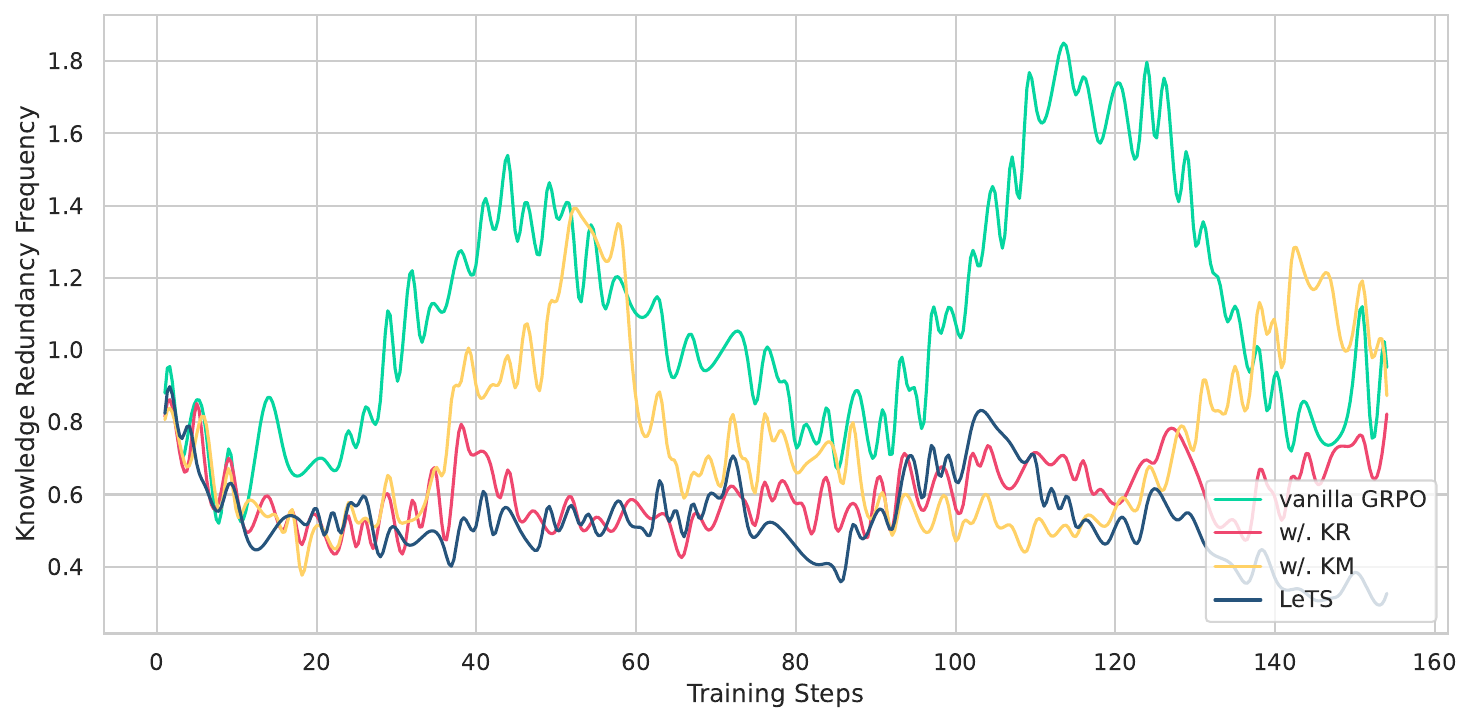}
  \caption{Knowledge redundancy frequency of the four methods during training.}
  \label{fig:training_kr}
\end{figure}

We further analyze the evolution of knowledge redundancy frequency across the four methods. Here, knowledge redundancy frequency refers to the rate at which identical documents are retrieved multiple times within a single rollout. As shown in Figure~\ref{fig:training_kr}, models trained with the in-rollout knowledge redundancy reward (\emph{i.e.}, w/. KR and \ours{}) consistently exhibit lower redundancy frequencies.

\subsection{Alleviation of the Redundant Searching Phenomenon}
To assess the impact of redundant searching on the performance of retrieval-augmented language models, we analyze the average frequency of redundant document retrieval. As shown in Table~\ref{tab:redudant}, both baseline models tend to retrieve more redundant information when they ultimately fail to answer questions correctly. In contrast, our proposed \ours{} significantly reduces in-rollout knowledge redundancy across four multi-hop benchmarks. These findings are consistent with the intended effect of the in-rollout knowledge redundancy reward.

\begin{table}[h]
\centering
\resizebox{\columnwidth}{!}{%
\begin{tabular}{@{}cc|cccc@{}}
\toprule
\multicolumn{2}{c|}{Dataset}        & HotpotQA & 2Wiki & MusiQue & Bamboogle \\ \midrule
\multirow{2}{*}{Correct} & ReSearch & 2.39     & 3.52  & 1.58    & 2.40      \\
                         & LeTS     & 0.23     & 0.37  & 0.21    & 0.47      \\ \midrule
\multirow{2}{*}{Incorrect}   & ReSearch & 3.53     & 4.49  & 2.94    & 3.21      \\
                         & LeTS     & 0.50     & 0.61  & 0.32    & 0.62      \\ \bottomrule
\end{tabular}%
}
\caption{Average in-rollout knowledge redundancy frequency of the two methods. We separately list the knowledge redundancy frequency of both correct and incorrect generations.}
\label{tab:redudant}
\end{table}

\subsection{Alleviation of the Irrelevant Searching Phenomenon}
To further validate the effectiveness of our approach in mitigating irrelevant searching, we evaluate the relevance of intermediate search queries generated by \ours{} compared to other methods. Appendix~\ref{appendix:case} presents two case studies that qualitatively compare our method with prior approaches. As shown in Table~\ref{tab:case}, \ours{} consistently produces more concise and relevant search queries.

\subsection{Efficiency Analysis}
We further evaluate the inference efficiency of our method on four multi-hop datasets. As shown in Table~\ref{tab:efficiency}, \ours{} substantially reduces both the number of generated tokens and the number of search steps during inference.
In particular, \ours{} demonstrates higher efficiency compared to prior methods by generating fewer tokens throughout the reasoning process. In terms of average search times, \ours{} achieves a reduction of approximately $30\%$, without compromising performance across all four benchmarks. These results suggest that our reinforcement learning framework—built on process- and outcome-level reward hybridization—more effectively guides language models to learn efficient think-and-search behavior.

\section{Conclusion}
In this paper, we first reveal two types of drawbacks in the current outcome-supervision RL for RAG. To address these points, we design two kinds of process-level rewards. Subsequently, we propose \ours{}, a novel RL framework that hybridizes the process- and outcome-based reward via an advantage rescaling approach. We further present the generalization and inference efficiency of our methods under different language models and numerous single-hop and multi-hop RAG benchmarks.

\section*{Limitations}
Despite \ours{} achieving significant results on the Qwen-7B and Qwen-3B series, due to cost constraints, we have not been able to validate our approach on larger language models.
Overall, our research empirically showcases the superiority of introducing the process-level supervision to outcome-based GRPO.

\bibliography{main}
\appendix

\section{Experimental Settings}
\label{appendix:exp}
\paragraph{Evaluation Benchmarks}
For evaluation on multi-hop scenarios, we consider four benchmark datasets: HotpotQA~\cite{yang-etal-2018-hotpotqa}, 2WikiMultiHopQA~\cite{ho-etal-2020-constructing}, MusiQue~\cite{trivedi-etal-2022-musique}, and Bamboogle~\cite{press-etal-2023-measuring}. Evaluation is conducted on the development sets of HotpotQA, 2WikiMultiHopQA, and MusiQue, and on the test set of Bamboogle, which contain 7405, 12576, 2417, and 125 samples, respectively. We also evaluate our method on three single-hop question answering datasets: NQ~\cite{kwiatkowski-etal-2019-natural}, PopQA~\cite{popqa}, and TriviaQA~\cite{joshi-etal-2017-triviaqa}.

\paragraph{Evaluation Metrics}
We evaluate our method using the commonly adopted Exact Match (\emph{EM}) metric, which considers a prediction correct only if it exactly matches the ground truth answer. To account for the overly strict nature of exact match in real-world scenarios, we additionally employ LLM-as-a-Judge (\emph{LJ}) for automatic evaluation. In this setting, \texttt{gpt-4o-mini} is used to assess the semantic correctness of the final answer. The prompt used for \emph{LJ} is provided in Appendix~\ref{appendix:lj_prompt}.

\section{Implementation Details}
\label{appendix:implementation_details}

Our training is conducted on 8 Nvidia A100 GPUs, with full parameter optimization and gradient checkpointing. The detailed hyperparameter settings are listed in Table \ref{tab:implementation-details}.


\begin{table}[htbp]
\begin{center}
\begin{tabular}{@{}l|l@{}}
\toprule[1.5pt]
\textbf{Parameter}          & \textbf{Value} \\ \midrule
Learning Rate               & 1e-6           \\
Train Batch Size            & 256            \\
Number of Training Epochs   & 2              \\
Number of Rollouts          & 5              \\
Rollout Temperature         & 1.0            \\
KL Loss Coefficient         & 0.001          \\
Clip Ratio                  & 0.2            \\
Num. Document Retrieval     & 5              \\
Rescale Factor $\lambda$    & 0.1            \\ \bottomrule[1.5pt]
\end{tabular}
\end{center}
\caption{Implementation details of \ours{}.}
\label{tab:implementation-details}
\end{table}


\section{Prompt Templates}
In this appendix, we provide detailed prompt templates used in various stages of our experiments. These templates ensure the consistency and reproducibility of our experimental setup.

\subsection{Prompt Templates for Training}
\label{appendix:training_prompt}
The training prompt templates are designed to clearly guide the language models in performing structured reasoning and retrieval tasks under both base and instruction-tuned settings. Specifically, Table~\ref{tab:training_prompt_base} illustrates the prompt used for base models, which frames the interaction as a conversation and explicitly defines the tags for reasoning, searching, and answering. Table~\ref{tab:training_prompt_ins} shows the system prompt for instruction-tuned models, where the model is guided to follow a similar reasoning-and-search pattern with clear structural conventions. These prompts are essential to elicit the think-and-search behavior in a controlled and consistent manner across training scenarios.

\vspace{1em}
\begin{table}[h]
\vspace{-1em}
\begin{prompt}[title={Prompt Template For Base Model}, label=prompt:training_prompt_base]
A conversation between User and Assistant. \
The user asks a question, and the assistant solves it. \
The assistant first thinks about the reasoning process in the mind and then provides the user with the answer. \
During thinking, the assistant can invoke the wikipedia search tool to search for fact information about specific topics if needed. \
The reasoning process and answer are enclosed within \texttt{<think>} \texttt{</think>} and \texttt{<answer>} \texttt{</answer>} tags respectively, \
and the search query and result are enclosed within \texttt{<search>} \texttt{</search>} and \texttt{<result>} \texttt{</result>} tags respectively. \
For example, \texttt{<think>} This is the reasoning process. \texttt{</think>} \texttt{<search>} search query here \texttt{</search>} \texttt{<result>} search result here \texttt{</result>} \
\texttt{<think>} This is the reasoning process. \texttt{</think>} \texttt{<answer>} The final answer is \verb|\boxed{answer here}| \texttt{</answer>}. \
In the last part of the answer, the final exact answer is enclosed within \verb|\boxed{}| with latex format. \
User: \textcolor{red}{\texttt{prompt}}. Assistant: 
\end{prompt}
\caption{Prompt Template For Base Model.\label{tab:training_prompt_base}}
\end{table}

\begin{table}[]
\vspace{-1em}
\begin{prompt}[title={System Prompt Template For Instruction-Tuned Model}, label=prompt:training_prompt_ins]
You are a helpful assistant that can solve the given question step by step with the help of the wikipedia search tool. \
Given a question, you need to first think about the reasoning process in the mind and then provide the answer. \
During thinking, you can invoke the wikipedia search tool to search for fact information about specific topics if needed. \
The reasoning process and answer are enclosed within \texttt{<think>} \texttt{</think>} and \texttt{<answer>} \texttt{</answer>} tags respectively, \
and the search query and result are enclosed within \texttt{<search>} \texttt{</search>} and \texttt{<result>} \texttt{</result>} tags respectively. \
For example, \texttt{<think>} This is the reasoning process. \texttt{</think>} \texttt{<search>} search query here \texttt{</search>} \texttt{<result>} search result here \texttt{</result>} \
\texttt{<think>} This is the reasoning process. \texttt{</think>} \texttt{<answer>} The final answer is \verb|\boxed{answer here}| \texttt{</answer>}. \
In the last part of the answer, the final exact answer is enclosed within \verb|\boxed{}| with latex format.
\end{prompt}
\caption{System Prompt Template For Instruction-Tuned Model.\label{tab:training_prompt_ins}}
\end{table}

\subsection{Prompt Templates for LLM-as-a-Judge Evaluation}
\label{appendix:lj_prompt}
The evaluation prompt in Table~\ref{tab:lj_prompt} defines the judgment criteria and expected output format for LLM-as-a-Judge, ensuring consistent and interpretable assessment of generated answers with respect to semantic correctness.

\vspace{1em}
\begin{table}[!ht]
\begin{prompt}[title={Prompt for LLM-as-a-Judge}, label=prompt:lj_prompt]
You will be given a question and its ground truth answer list, where each item can be a ground truth answer. Provided a pred\_answer, you need to judge if the pred\_answer correctly answers the question based on the ground truth answer list. You should first give your rationale for the judgment, and then give your judgment result (i.e., correct or incorrect).

Here is the criteria for the judgment:
1. The pred\_answer doesn't need to be exactly the same as any of the ground truth answers, but should be semantically the same for the question.
2. Each item in the ground truth answer list can be viewed as a ground truth answer for the question, and the pred\_answer should be semantically the same as at least one of them.

question: {question}

ground truth answers: {gt\_answer}

pred\_answer: {pred\_answer}

The output should be in the following json format:

```json 

\{

    "rationale": "your rationale for the judgement, as a text",
    
    "judgement": "your judgement result, can only be `correct' or `incorrect'"
    
\}

'''

Your output:
\end{prompt}
\caption{Prompt for LLM-as-a-Judge.\label{tab:lj_prompt}}
\end{table}

\newpage
\section{Learning to Search via Reinforcement Learning on Base Version of Small Language Models}
\label{appendix:slm}
\begin{figure}[]
  \centering
  \includegraphics[width=\columnwidth]{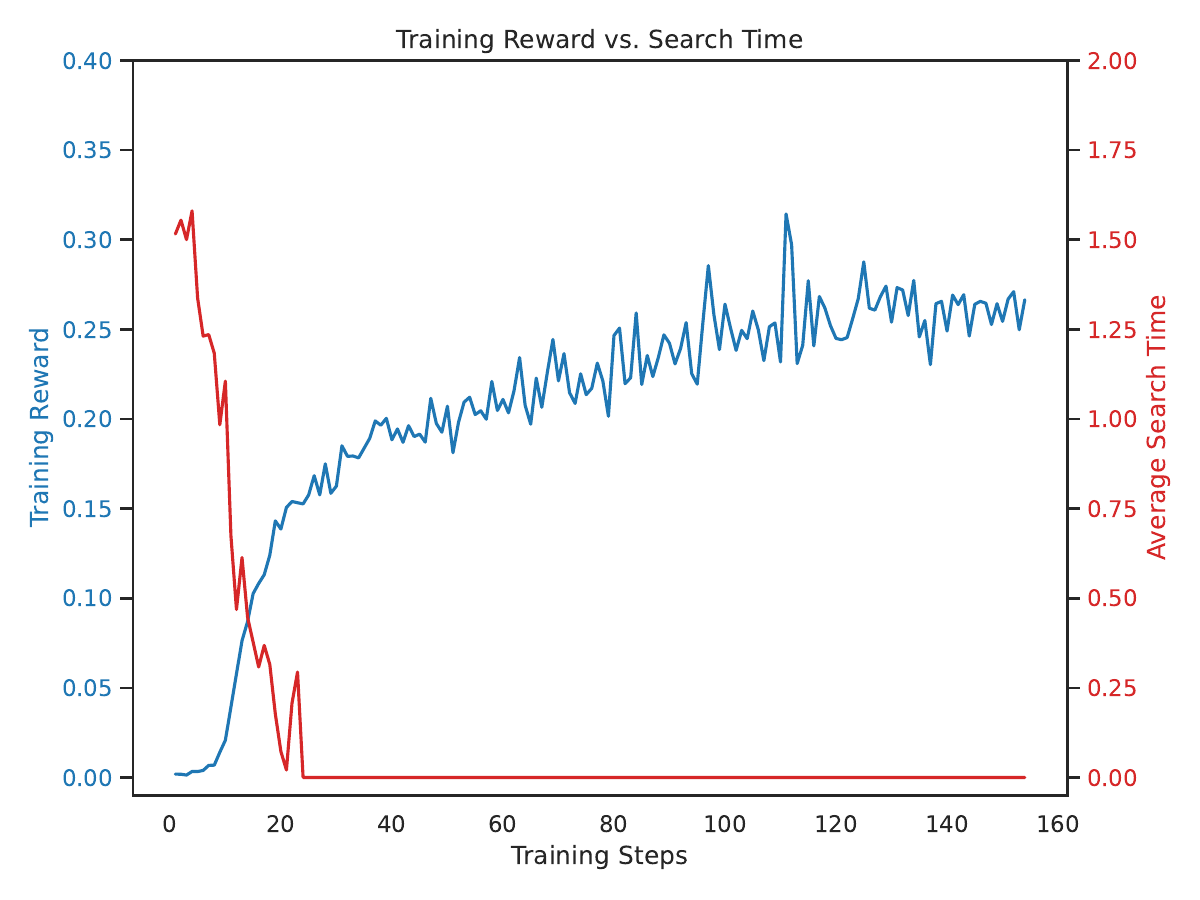}
  \caption{The variance of reward and average search time with Qwen2.5-3B-Base as the backbone during GRPO training.}
  \label{fig:slm}
\end{figure}

In this section, we further dive into learning to search via reinforcement learning on the base version of smaller language models (\emph{i.e.}, Qwen2.5-3B-Base). We start by reproducing ReSearch on Qwen2.5-3B-Base with the same settings as Table~\ref{tab:main_results}. As shown in figure~\ref{fig:slm}, as the reward keeps increasing to around $0.3$, the model's average search time soon decreases to $0$ due to reward hacking.

Therefore, we can conclude that the current F1 score-based outcome-level reward fails to stimulate the think-and-search ability of the base version of smaller language models. We leave this as an open problem for subsequent research.

\newpage
\section{Pseudo Code for \ours{}}
The pseudo code for \ours{} is shown in Algorithm~\ref{appendix:pseudo}.
\label{appendix:pseudo}
\begin{algorithm}[]
\caption{Process-and-Outcome Reward Hybridized GRPO in \ours{}}
\begin{algorithmic}[1]
\Require Policy model $\pi_{\theta_{\text{old}}}$, reference model $\pi_{\theta_{\text{ref}}}$, dataset $\mathcal{D}$, rescale factor $\lambda$, KL penalty $\beta$
\For{each input $x \sim \mathcal{D}$}
    \State Sample $G$ rollouts $\{y_i\}_{i=1}^G \sim \pi_{\theta_{\text{old}}}(x)$
    \For{each rollout $y_i$}
        \State Compute outcome-level reward $r^{(o)}_i$ via Eq.~\ref{eq:outcome}
        \If{rollout is well-formatted}
            \If{$y_i$ is in outperforming group}
                \State Compute stepwise process rewards $r^{(p)}_{i,j}$ using Eq.~\ref{eq:sim}
            \Else
                \State Compute $r^{(p)}_{i,j}$ via group match reward using Eq.~\ref{eq:match}
            \EndIf
        \EndIf
        \State Normalize $\{r^{(o)}_i\}_{i=1}^G$ to get $A^{(o)}_i$
        \State Normalize $\{r^{(p)}_{i,j}\}_{j=1}^{n_i}$ to get $\hat{r}^{(p)}_{i,j}$
        \For{each step $j$}
            \State $A_{i,j} \gets \left( 1 + \mathrm{sgn}(A^{(o)}_i) \cdot \lambda \hat{r}^{(p)}_{i,j} \right) \cdot A^{(o)}_i$ 
            \Comment{Eq.~\ref{eq:rescale}}
            \State Compute GRPO loss using $A_{i,j}$ via Eq.~\ref{eq:grpo}
        \EndFor
    \EndFor

\EndFor
\end{algorithmic}
\end{algorithm}

\newpage
\section{Case Studies}
\label{appendix:case}
In this section, we present comparative case studies derived from the interaction trajectories of \textbf{ReSearch} and our proposed \textbf{\ours{}}. Specifically, Table~\ref{tab:case} illustrates a scenario where \textbf{ReSearch} incorrectly answered a multi-hop question due to redundant and irrelevant search steps, whereas our proposed \textbf{\ours{}} successfully provided the correct answer. This clearly demonstrates the effectiveness of integrating process-level rewards with outcome-level supervision.

\begin{table*}[t]
\centering
\footnotesize
\begin{tabular}{p{13.5cm}}
\toprule
\textbf{Question}: When did the president who set the precedent of a two term limit enter office? \\
\textbf{Golden Answer}: April 30, 1789 \\
\midrule

\textbf{ReSearch} \textcolor{red}{(Incorrect)}: Misled by irrelevant searching leading to incorrect answer. \\
\think{To determine when the president who set the precedent of a two-term limit entered office, I need to first identify which president this is. I will use the Wikipedia search tool to find this information.}

\search{\colorbox{green!20}{Who was the first president to set the two-term limit precedent?}}

\info{"Term limits in the United States"
Republican government as a periodic rotation"". The historian Mercy Otis Warren, warned that ""there is no provision for a rotation, nor anything to prevent the perpetuity of office in the same hands for life; which by a little well-timed bribery, will probably be done"". Michael Korzi says ...}

\think{ Now that I have the information, I can identify the president who set the precedent of a two-term limit. It was Thomas Jefferson. Now I need to find out when he entered office. }

\search{\colorbox{red!20}{When did Thomas Jefferson enter office?}}

\info{"Presidency of Thomas Jefferson"
Presidency of Thomas Jefferson The presidency of Thomas Jefferson began on March 4, 1801, when he was inaugurated as the third President of the United States, and ended on March 4, 1809. Jefferson assumed the office after defeating incumbent President John Adams in the 1800 presidential election. The election was ...} 
\textcolor{red}{(Irrelevant Document)}

\think{Now I have the information I need. Thomas Jefferson entered office on March 4, 1801.}

\search{\colorbox{red!20}{When did Thomas Jefferson become president?}}

\info{"Presidency of Thomas Jefferson"
Presidency of Thomas Jefferson The presidency of Thomas Jefferson began on March 4, 1801, when he was inaugurated as the third President of the United States, and ended on March 4, 1809. Jefferson assumed the office after defeating incumbent President John Adams in the 1800 presidential election. The election was ...} 
\textcolor{red}{(Irrelevant and Redundant Documents)}

\think{After reviewing the information, I can now provide the answer to the question. Thomas Jefferson, the president who set the precedent of a two-term limit, entered office on March 4, 1801.}

\answer{\boxed{March~4,~1801}}\\
\midrule

\textbf{\ours{}} \textcolor{green}{(Correct)}: Effective, concise search leading to the correct answer. \\
\think{First, I need to identify which president set the precedent of a two-term limit. Then, I need to find out when that president entered office. Let's start by finding out which president established this precedent.}

\search{\colorbox{green!20}{which president established the two-term limit precedent}}

\info{"Twenty-second Amendment to the United States Constitution"
Twenty-second Amendment to the United States Constitution The Twenty-second Amendment (Amendment XXII) to the United States Constitution sets a limit on the number of times an individual is eligible for election to the office of President of the United States, and also sets additional eligibility conditions for presidents who succeed to the unexpired terms of their predecessors ...}

\think{ From the information provided, it is clear that Franklin D. Roosevelt was the first president to break the two-term tradition by winning a third and fourth term. The precedent of a two-term limit was established by George Washington, who retired after two terms. The next step is to find out when George Washington entered office.
}

\search{\colorbox{green!20}{when did george washington enter office}}

\info{"Presidency of George Washington"
Presidency of George Washington The presidency of George Washington began on April 30, 1789, when Washington was inaugurated as the first President of the United States, and ended on March 4, 1797. Washington took office after the 1788–89 presidential election, the nation's first quadrennial presidential election, in which he was ...}

\think{The information clearly states that George Washington was inaugurated as the first President of the United States on April 30, 1789. This is the date when he entered office.}

\answer{\boxed{April~30,~1789}}\\

\bottomrule
\end{tabular}
\caption{Case study on the irrelevant searching phenomenon.}
\label{tab:case}
\end{table*}



\end{CJK}
\end{document}